\documentclass{article}
\usepackage{jfrExamplee}
\usepackage{graphicx}
\usepackage{apalike}
\usepackage{setspace}
\usepackage{sidecap}
\usepackage{array}
\usepackage{amsmath}
\usepackage{hyperref}

\title{KiloBot: A Programming Language for Deploying\\ Perception-Guided Industrial Manipulators at Scale}

\author{
Wei Gao\thanks{Corresponding author. Email: gaowei19951004@hotmail.com} \\
Mech-Mind Robotics\\
\And
Jingqiang Wang \\
Mech-Mind Robotics \\
\AND
Xinv Zhu \\
Mech-Mind Robotic \\
\And
Jun Zhong \\
Mech-Mind Robotic \\
\And
Yue Shen \\
Mech-Mind Robotic \\
\And
Youshuang Ding \\
Mech-Mind Robotic \\
}

\begin{document}

\maketitle

\begin{abstract}
We would like industrial robots to handle unstructured environments with cameras and perception pipelines. In contrast to traditional industrial robots that replay offline-crafted trajectories, online behavior planning is required for these perception-guided industrial applications.
Aside from perception and planning algorithms, deploying perception-guided manipulators also requires substantial effort in integration.
One approach is writing scripts in a traditional language (such as Python) to construct the planning problem and perform integration with other algorithmic modules \& external devices. While scripting in Python is feasible for a handful of robots and applications, deploying perception-guided manipulation at scale (e.g., more than 10000 robot workstations in over 2000 customer sites) becomes intractable.
To resolve this challenge, we propose a Domain-Specific Language (DSL) for perception-guided manipulation applications. To scale up the deployment, our DSL provides: 1) an easily accessible interface to construct \& solve a sub-class of Task and Motion Planning (TAMP) problems that are important in practical applications; and 2) a mechanism to implement flexible control flow to perform integration and address customized requirements of distinct industrial application.
Combined with an intuitive graphical programming frontend (Figure.~\ref{fig:frontend}), our DSL is mainly used by machine operators without coding experience in traditional programming languages. Within hours of training, operators are capable of orchestrating interesting sophisticated manipulation behaviors with our DSL.
Extensive practical deployments demonstrate the efficacy of our method.
\end{abstract}

\section{Introduction}

The wide availability of RGBD cameras provides robots with powerful 3D sensing capabilities. As a result, robots equipped with these sensors are entering industrial production to handle unstructured environments.
As the task environment is not static, and manipulated objects in these applications are perceived from cameras, the robot behaviors must be planned online (instead of crafted offline). This type of behavior planning problem contains elements of discrete decision-making and continuous motion generation, and is denoted as Task and Motion Planning (TAMP). Extensive contributions~\cite{alami1990geometrical,hauser2010multi,garrett2018sampling,srivastava2014combined} have been made regarding this topic and many open-source packages are available. Please refer to~\cite{garrett2021integrated} for a detailed review.

Despite these excellent contributions, deploying perception-guided manipulators requires substantial effort in integration. This is the procedure of: 1) constructing the TAMP problem as the input to the planner; 2) putting different modules (e.g., perception) together; and 3) implementing control flows to address customized requirements of industrial applications. Writing integration scripts in Python is feasible for a handful of applications.
However, planning problems and control flows in these scripts are tightly coupled with field works such as hardware setup, tuning for algorithm parameters \&\& movement targets, sensor calibration, and communication with external devices. Thus, the integration code among different applications is almost non-reusable. Consequently, close collaboration between a programmer and a field application specialist is required for deploying each application, which is time-consuming and expensive.

\begin{figure}[t]
    \centering
    \includegraphics[width=0.7\textwidth]{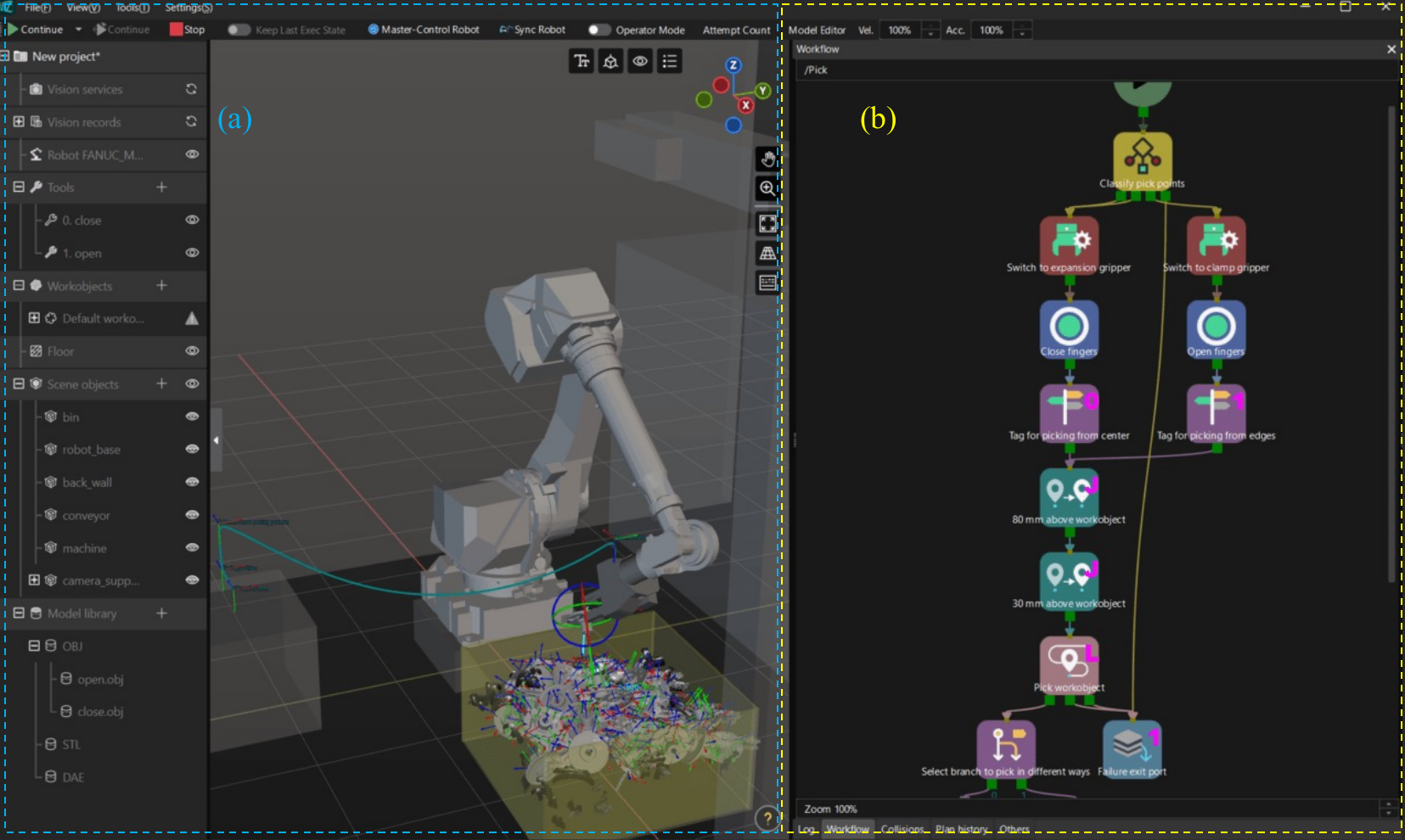}
    \caption{The graphical programming frontend for our programming language. Users orchestrate the robot behaviors by drag-and-drop programming to construct a control-flow graph, as shown in region (b) on the right. The region (a) on the left is the 3D visualization and resource window. }
    \label{fig:frontend}
\end{figure}

The ``scripting" approach mentioned above in a traditional programming language has limited scalability. Thus, we propose a new Domain-Specific Language (DSL) to reduce the integration effort and scale up the deployment. Regarding the front end, we design an intuitive graphical programming interface mainly for field application engineers or machine operators without experience in traditional programming languages (Python/C++). This is inspired by many existing graphical programming languages~\cite{burnett1995visual,tsai2019improving} intended for education and entertainment that target people without coding experience. As shown in Figure.~\ref{fig:frontend}, users can orchestrate the robot behaviors by drag-and-drop programming to construct a \textit{control-flow graph} (detailed in Sec.~\ref{sec:preliminary}).

The backend of our DSL is responsible for running the TAMP algorithm and executing the user-defined control flow. We propose a novel interface between the DSL and the planning algorithm: users only need to craft a ``skeleton'' of the desired robot behavior, where the skeleton might contain a set of parameters (both discrete and continuous) not determined offline. Then, the planning algorithm generates these missing parameters during online execution. In this interface, users do not need to understand the TAMP algorithm or explicitly implement the planning problem description (in a modeling language like PDDL). Thus, the interface is user-friendly even for machine operators without coding experience.

The contributions of this paper are as follows: 1) we design a novel DSL for machine operators without coding experience to deploy perception-guided robot manipulation applications. 2) We propose a novel interface that implicitly integrates a customized TAMP 
problem description and planning algorithm into our DSL. The interface is user-friendly, and the TAMP planner meets the performance requirement for deployment. 3) Inspired by pipelining in modern CPU, we introduce a ``pre-planning'' interpreter for our DSL that interleaves robot movement with planning (for future robot behaviors). This pre-planning mechanism significantly improves the manipulator's throughput (\# of pick-place per minute). 4) We conduct extensive tests of our DSL during the deployment of more than 10000 robot workstations worldwide. Please visit \href{https://www.mech-mind.com/}{\underline{https://www.mech-mind.com/}} for more examples.

This paper is organized as follows: in Sec.~\ref{sec:related} we introduce related works. Sec.~\ref{sec:preliminary} presents the preliminaries. Sec.~\ref{sec:interface} presents the design of our DSL and the interface with TAMP planner. Sec.~\ref{sec:interpreter} introduce the implementation of interpreter and the pre-planning mechanism. Sec.~\ref{sec:result} shows the results. Sec.~\ref{sec:conclusion} concludes.

\section{Related Work}
\label{sec:related}

\subsection{Robot Offline Programming Software}
\label{subsec:offline}

Robot offline programming~\cite{wittenberg1995developments,pan2012recent} means generating robot programs in a virtual environment based on 3D CAD data (instead of pendant teaching). Offline programming software is typically equipped with advanced collision detection and motion planning algorithms to generate robot movement for various industrial applications, such as welding, coating, dispensing, and robot milling. Users can inspect the generated movement in the integrated simulator of the software. Once the movement is verified, it can be downloaded to the physical robot for online execution.

Offline programming softwares~\cite{robotdk,robotstudio,robotmaster} have been extensively deployed in practice and they accomplish many challenging manipulation tasks. However, the offline-generated robot movements cannot adapt to dynamic or unstructured working environments, where the manipulated objects must be perceived online. Our system is proposed to resolve this limitation using online perception and planning pipelines.

\subsection{Integrated Task and Motion Planning}

As described in~\cite{garrett2021integrated}, task and motion planning (TAMP) is the problem of finding actions of a robot that moves itself and changes the state of the environment objects. TAMP contains elements of discrete task planning and continuous motion planning. Extensive contributions~\cite{alami1990geometrical,hauser2010multi,garrett2018sampling,srivastava2014combined} have been made on strategies to solve TAMP problems. Many of these algorithms require an internal planner to solve the joint-space collision-free motion planning problem. The most effective methods are based on sampling~\cite{kavraki1996probabilistic,lavalle2001randomized} and trajectory optimization~\cite{ratliff2009chomp,schulman2014motion}.

Our work is built upon these excellent contributions regarding the formulation and planning algorithms of TAMP problems. Actually, one prominent feature of our DSL is to serve as an interface layer that converts user control-flow graph (with undetermined parameters) into a series of problem descriptions that can be solved by existing TAMP algorithm~\cite{garrett2018sampling}.

%
%

\subsection{Manipulation Pipelines}

Researchers have created robot manipulation pipelines~\cite{zeng2021transporter,florence2019self,gao2021kpam,xia2022review} with interesting capabilities. These methods typically integrated various perception and planning modules to achieve intelligent manipulation behaviors. Some pipelines accept inputs from other modality, such as language~\cite{liang2023code,driess2023palm} or tactile sensors~\cite{maldonado2012improving,su2023customizing}.
Compared to these excellent works, our DSL is designed to address different challenges. The manipulation behavior programmed by our DSL is typically much less innovative than these works. On the other hand, we would like to achieve deployment at scale by reducing the integration cost and address customized requirements in industrial applications that are diverse, application-specific and tightly coupled with field work. 

\section{Preliminary}
\label{sec:preliminary}

\begin{figure}[t]
    \centering
    \includegraphics[width=0.75\textwidth]{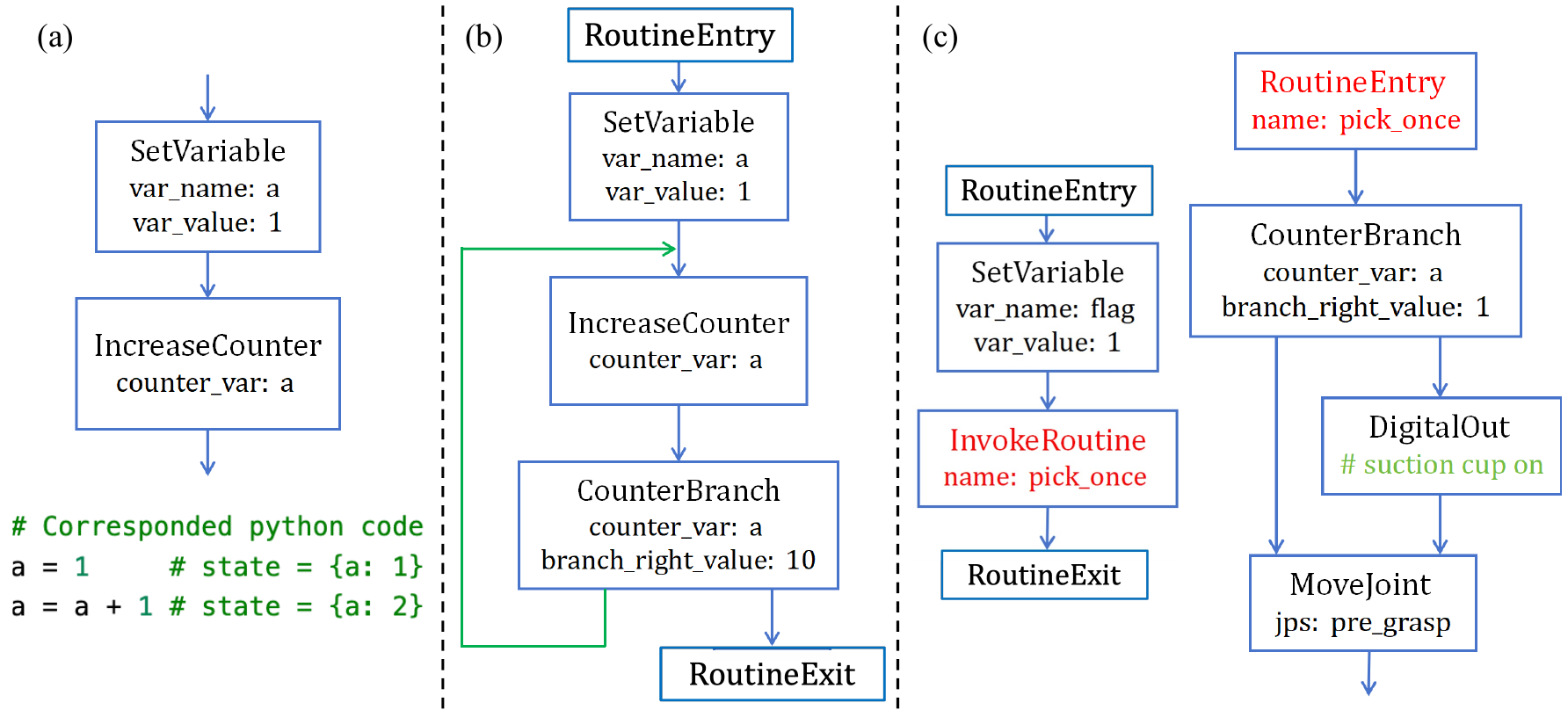}
    \caption{Examples of control flow graph representation of user programs. (a) Nodes in a control flow graph correspond to a Python \textit{statements}. (b) Edges in a control flow graph can be used to implement branch and loop structures, such as the green edge. (c) Similar to functions in Python, nodes in a control flow graph can be organized into \textit{routines}, which can be invoked from another routine. }
    \label{fig:cfg}
\end{figure}

\subsection{Control Flow Graph}
\label{subsec:cfg}

Our DSL, as well as several existing programming languages with graphical frontends, enables users to explicitly construct control flow graphs by drag-and-drop operations. In this sub-section, we present an overview of the control flow graph in this context and compare it with Python, a traditional interpreted programming language.

A control flow graph is a ``node and edge'' representation of the user program, an example is shown in Figure.~\ref{fig:cfg}. Each node in the control flow graph roughly corresponds to a \textit{statement} in Python. The statement can perform read and/or write operations to the variable map (environment), which is a map from variable name string to variable value. A statement can also produce side effects, such as writing a file or sending a message. In the following text, we would use \texttt{node\_name(user\_parameters)} to denote nodes, for example \texttt{IncreaseCounter(counter\_var:str)}. We omit the \texttt{(user\_parameters)} when the context is clear.

A directed edge in a control flow graph implies the execution order of different nodes, which roughly corresponds to the \texttt{goto} statement in Python. Thus, edges in a control flow graph can be used to implement branch and loop structures, which corresponds to \texttt{if/for/while/break/continue} in Python. An example is shown in Figure~\ref{fig:cfg} (b), where the directed edge implements a loop.

Similar to Python, nodes in a control flow graph can be organized into \textit{routines} (or \textit{functions}). A routine is also a node graph that can be \textit{invoked} by another routine. An example is shown in Figure~\ref{fig:cfg} (c). A routine has exactly one entry, which is the \texttt{RoutineEntry} node. A routine has one or more exits, which is marked by \texttt{RoutineExit} nodes. One particular routine, denoted as the main routine, is the global entry of the entire program. Except for the \texttt{RoutineEntry} node, every node has exactly one in-port. Similarly, every node has one or more out-port except for \texttt{RoutineExit} node.

\subsection{Specialization for Pick-and-Place Manipulation Applications}
\label{subsec:specialization}

\begin{table}[t]
\centering
\begin{tabular}{ |m{4.3cm}|m{10.5cm}|}
\hline
Variable Name: Type & Description  \\
\hline
\texttt{jps}: \texttt{Vector}& Robot joint-space configuration. \\
\hline
\texttt{active\_tool}: \texttt{Int} & Index of the currently used gripper tool. \\
\hline
\texttt{static\_env}: \texttt{Compound} & Data that are usually not mutated during execution, such as robot model, static collision geometry, gripper tool setup. \\
\hline
\texttt{picked\_objects}: \texttt{Compound} & Objects that are picked by the robot. The geometry, pose wrt gripper, and other meta-info of each object is included. \\
\hline
\texttt{placed\_objects}: \texttt{Compound} & Objects that have been placed by the robot. The geometry, pose in world, and other meta-info of each object is included. \\
\hline
\texttt{\{srvID\}\_perception}: \texttt{Json} & RPC response from the perception service identified by \texttt{srvID}. It contains pose, geometry, grasping and other info about detected objects.  \\
\hline
\end{tabular}
\caption{Several important variables used in our DSL. }
\label{table:table_vars}
\end{table}

Our DSL can be regarded as a control flow graph with interface for constructing and solving TAMP problems. This interface is discussed in Sec.~\ref{sec:interface}. In this subsection, we present several design decisions not directly related to planning, which serves as the background for further discussion.

\vspace{1mm}
\noindent \textbf{Frontend and backend:}
Our implementation of the DSL is separated into the frontend and backend, and both of them are represented as control flow graphs. The frontend, illustrated in Figure.~\ref{fig:frontend}, is designed for user-friendliness with a lot of ``language sugar''. The frontend control graph, as a serializable representation of the user program, can be converted into a backend control flow graph for execution. The discussion in the following text mainly focuses on the backend.

\vspace{1mm}
\noindent \textbf{Variable mechanism:} Table.~\ref{table:table_vars} summarizes several important variables in our DSL. In addition to these, user can define their own variables (e.g., by~\texttt{SetVariable} node) and make arbitrary mutations to them. The backend provides a \texttt{FunctorVariableMutation} node, which contains a pure C++ functor (\texttt{std::function}) that takes the variable map as input and produces a list of mutations to that variable map. Nearly all frontend nodes that are unrelated to planning are converted into this node in the backend, such as the \texttt{IncreaseCounter} in Figure.~\ref{fig:cfg}. Moreover, advanced users that are capable of programming can implement their own fuctors and insert them into the control flow graph by this node, which is used to address very complex application requirements.

In our DSL, all variables are \textit{global}. In particular, there are no local variables for a routine. The information exchange between the routine caller and callee is achieved by reading and/or writing of global variables. This design decision is made because our DSL does not aim at complex control flows. Sophisticated algorithms and operations are either embedded in the behavior planner or provided as pre-defined nodes for users to drag-and-drop.

\vspace{1mm}
\noindent \textbf{External communication:} Our DSL communicates with external algorithmic modules and devices through Remote Procedure Call (RPC). Our discussion would be restricted to synchronized RPC for simplicity, while asynchronous RPC is used by default in the backend for efficiency.

A node \texttt{CallService(srvID:str, request\_var:str, response\_save\_var:str)} can be used by programmers to invoke a RPC service. This node has the following behavior: 1) find the service from registered ones by the \texttt{srvID}; 2) Pack and send the request message, which contains meta info (e.g., timestamp and message ID) and optionally a serializable variable identified by \texttt{request\_var}; and 3) wait for the response message synchronously (blockingly), and save the response message to a variable named \texttt{response\_save\_var}.

For example, the entire perception stack is an RPC service in our pipeline. This includes invoking the camera to take an image, running a series of perception algorithms (object detection, pose estimation, occlusion detection, grasping pose generation), and sending the result back to the RPC caller. By default, the response variable of perception service is \texttt{\{srvID\}\_perception}. RPC is also used to communicate with other algorithmic modules (e.g., palletization pattern generation) and external devices (e.g., conveyors).




\section{Interface with Robot Behavior Planning}
\label{sec:interface}

A TAMP algorithm is integrated into our DSL to alleviate the users' burden of making discrete decisions and crafting robot trajectories. We propose the following interface between the language and the planner: some pre-defined nodes are used to specify a ``skeleton'' of desired robot behavior, and they are intentionally undetermined offline. The planner converts these skeleton nodes into executable nodes by providing them with a set of discrete and/or continuous parameters. For notational clearance, we use \textit{online-parameters} for a given node to denote its parameters generated by behavior planning. This is in contrast to user-parameters of nodes mentioned in Sec.~\ref{subsec:cfg}, which are explicitly provided by the users.

For example, the \texttt{IncreaseCounter(counter\_var:str)} node has one user-parameter: a string indicating the counter variable name to be increased. It does not need an online-parameter as it is not involved with planning. On the other hand, for movement nodes (detailed in Subsec.~\ref{subsec:move_nodes}), online parameters are the planned joint-space trajectories and the safety certificate of the trajectories.
It is emphasized that online-parameters are not visible to users. They are part of the runtime data used to execute nodes that need planning. Thus, to make the DSL user-friendly, we only need to simplify the user-parameters of nodes.

In the following subsections, we describe this interface in detail. Subsec.~\ref{subsec:example} gives an overview with an illustrative example. Subsec.~\ref{subsec:move_nodes} and Subsec.~\ref{subsec:pick_node} present nodes for movement and pick-place behaviors, respectively. Subsec.~\ref{subsec:pr_struct} describes the integration of planning into the control flow of our DSL.

\subsection{Illustrative Example}
\label{subsec:example}

\begin{figure}[t]
    \centering
    \includegraphics[width=0.85\textwidth]{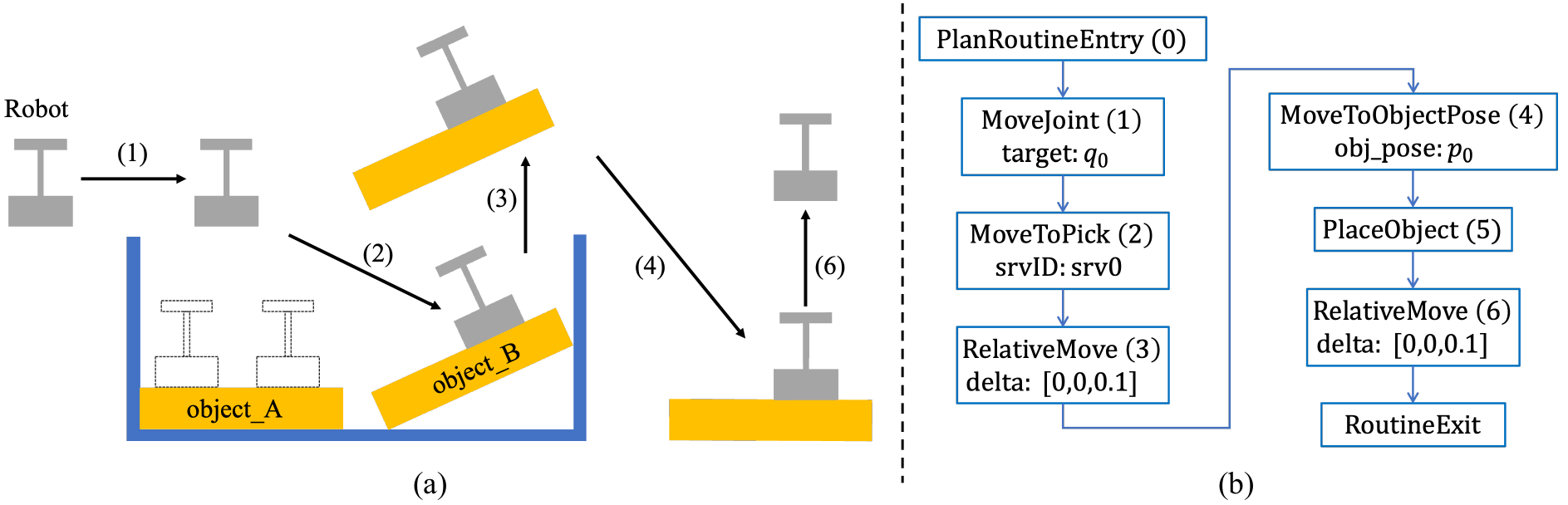}
    \caption{A schematic example of robot pick-and-place behaviors and the corresponding user program in our DSL. Perception service provides a scene with two objects (A and B), each with two possible grasping poses (in dash lines), as shown in (a). The user program, as shown in (b), contains nodes that are intentionally undetermined offline and can not fully specify the robot behaviors. During execution, the interpreter invokes the behavior planning to make discrete decisions (e.g., selection of object instance and grasping pose) and generates robot trajectories. If the planning succeeds, the interpreter executes the user program given the planning result (online-parameters of each node). Please refer to Subsec.~\ref{subsec:example} for a detailed explanation. }
    \label{fig:example}
\end{figure}

In this subsection, we present an overview of our DSL using a schematic example, as shown in Figure.~\ref{fig:example}. Suppose the perception service provides a scene with two objects (A and B), each object has two possible grasping poses (in dash lines), as shown in Figure.~\ref{fig:example} (a). The user program for a pick-place manipulation is shown in Figure.~\ref{fig:example} (b), with user-parameters annotated for each node.

The \texttt{MoveJoint} node (1) moves the robot from its initial configuration to a new configuration on top of the container. This node needs a target joint position as the user-parameter. It also requires a \texttt{trajectory\_config} user parameter to specify how the robot should reach its target (e.g., straight line or RRT-generated path). To provide a clear presentation, this user-parameter is omitted in Figuire~\ref{fig:example}.

The \texttt{MoveToPick} node (2) has the following behavior intuitively: 1) move the robot to a configuration that can grasp an object; and 2) pick up the object by attaching it to the robot end-effector. In the simplest form, the only user-parameter for this node is the name of the perception service response variable (\texttt{\{srvID\}\_perception} by default). The planner automatically figures out which object to pick, what is the optimal grasping pose and how to reach the robot-picking configuration as the online-parameters. Thus, users only need to provide high-level supervision, while the detailed decision-making and trajectory crafting are handled by the planner. Additional user-parameters can be used to guide the decision-making, as detailed in Subsec~\ref{subsec:pick_node}.

The \texttt{RelativeMove} node (3) and \texttt{MoveToObjectPose} node (4) are both movement nodes that move the robot to new configurations. The \texttt{RelativeMove} node (3) applies a constraint between the end-effector poses before and after this node, and it is used to lift the object in this example. The user-parameter of \texttt{RelativeMove} node is the relative transformation that defines the constraint. The \texttt{MoveToObjectPose} node (4) moves the robot to a joint configuration such that the picked object is at the user-specified pose. In its simplest form, this node only needs the object pose as the user-parameter. Alternatively, users can provide a map from object type to pose. Thus, this node would move the picked object to different poses according to its type.

The \texttt{PlaceObject} node (5) places the picked object(s) by detaching it from the robot end-effector and re-attaching it to the world. This node does not need an user-parameter or online-parameter. However, this node is involved in planning because it changes the geometry attachment and affects the collision checking of future movement nodes, such as the \texttt{RelativeMove} (6) in this example.

In this example, nodes (1)-(6) appear independent from each other. However, the underlying behavior planning must consider many nodes jointly, as these nodes' online parameters (both discrete or continuous ones) are coupled. For example, the selection of objects and grasping pose in \texttt{MoveToPick} (2) would affect the collision checking and trajectory generation of movement nodes (3-4) and \texttt{PlaceObject} node (5). An inappropriate picking decision, without considering the subsequent transferring and placement of the picked object(s), might lead to unavoidable collision or kinematic infeasibility.

To address this issue, we propose to give users the authority to specify nodes that must be planned jointly. In particular, a special type of routine, named \textit{plan-routine}, defines the scope of one behavior planning problem. The example in Figure.~\ref{fig:example} (b) is a plan-routine with a special \texttt{PlanRoutineEntry} node (0). With this formulation, the interpreter understands the discrete decision for \texttt{MoveToPick} (2) must consider node (1) and (3-6). In our practice, a plan-routine typically contains one iteration of pick-place operation. As the plan-routine is the basic unit of behavior planning, it cannot contain arbitrary topology structures (e.g., no loop). For now, we assume the plan-routine is a simple sequence. This constraint is relaxed in Subsec.~\ref{subsec:pr_struct}.

\subsection{Nodes for Robot Movement}
\label{subsec:move_nodes}

In this subsection, we describe nodes for robot movement. All movement nodes, for instance ones in Figure.~\ref{fig:example}, are defined by two generic user-parameters: the \texttt{target} of the movement and the \texttt{trajectory\_config}. Both user-parameters are used to generate robot trajectories during behavior planning, and they might induce various discrete decisions detailed below.

\begin{figure}[t]
    \centering
    \includegraphics[width=0.65\textwidth]{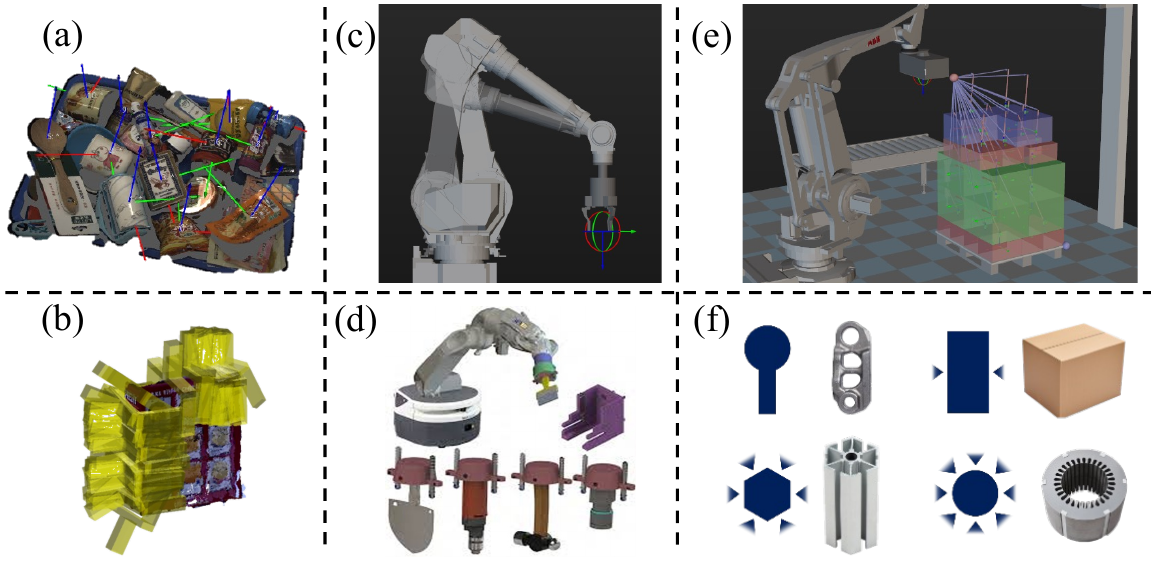}
    \caption{Representative examples of discrete decisions in our DSL. (a) Select the object to pick from a set of perceived objects. (b) Select the grasping pose of the object-to-pick. (c) Decide the best IK solution for a given end-effector pose. (d) Choose the gripper tool from multiple candidates. (e) Select the target from a target set, which can be either automatically generated (e.g., in palletization applications) or explicitly specified by the user. (f) Select the symmetry of the object for a given object pose target. }
    \label{fig:decision}
\end{figure}

The \texttt{target} user-parameter, provided by the human operator, specifies the high-level movement target, which would eventually be resolved as a fully determined joint target during planning. This high-level \texttt{target} might be provided in many forms, such as:

\begin{itemize}
    \item A robot joint target (\texttt{MoveJoint} node in Figure.~\ref{fig:example}). No decision-making is required for this target.
    \item A pose target for the robot end-effector. For this target, the inverse kinematics is invoked which generates several solutions in joint space (6-DoF robots typically have 8 solutions), as illustrated in Figure.~\ref{fig:decision} (c). The planner should evaluate the feasibility of these solutions and select the best one according to some metrics (e.g, minimum joint-space distance).
    \item A pose target for the picked object (\texttt{MoveToObjectPose} node in Figure.~\ref{fig:example}). One object pose target might be transformed into multiple end-effector pose targets due to the symmetry of the picked object(s), which occurs frequently in industrial applications. An illustration is shown in Figure.~\ref{fig:decision} (f). The planner should attempt these end-effector pose targets and select the best one.
    \item A (discrete or continuous) set of pose targets for picked object(s) or end-effector. The most prominent example is the \texttt{PalletizationMove} node, where the picked box(es) can be placed into multiple positions of a pallet. An illustration is shown in Figure.~\ref{fig:decision} (e). The planner might need to consider various factors for this decision, such as the feasibility of future palletization movement.
    \item A target that depends on the intermediate output of the planner. For instance, a \texttt{RelativeMove} node that depends on the previous target or \texttt{PalletizationMove} node that depends on the selected box(es) for picking. Generally, the movement target can be a function of previous/future movement targets, object properties, active tool and picking state. The planner should correctly resolve those dependency during planning.
\end{itemize}

On the other hand, the \texttt{trajectory\_config} user-parameter specify how should the robot reach its target. The trajectory might be a simple straight line in joint/end-effector space, a selection from a trajectory library, or a complex trajectory generated by an advanced motion planner (e.g., RRT). This \texttt{trajectory\_config} parameter also includes various user preference on the trajectory, such as collision option, movement speed configuration and singularity detection option. Moreover, a sequence \texttt{target} and \texttt{trajectory\_config} parameters can be received from RPC messages and decoded into a \texttt{robot\_trajectory} variable, which is the user-parameter of the \texttt{MoveTrajectoryByVariable} node. This enables the robot to execute movements from external commands.

All the movement nodes have the same types of online-parameter: the planned robot joint-space trajectories and the safety certificate of the trajectories. Given the online-parameters, the execution of movement nodes would: 1) send the planned joint-space trajectory to the robot service (which is an RPC service) for execution; and 2) update the \texttt{jps} variable (Table.~\ref{table:table_vars}) to the final joint configuration of the generated trajectory.

\subsection{Node for Object Picking}
\label{subsec:pick_node}

In this subsection, we formally describe the \texttt{MoveToPick} node introduced in Subsec.~\ref{subsec:example}, which plays a critical role in our DSL. For robot picking, the robot needs to use various types of gripper tools (suction cup, parallel-jaw, etc), move to appropriate joint-space configuration and pick up one or several objects. The perception pipeline produces objects available for pick, the method (gripper tool index, picking pose wrt objects, digital-out ports) to pick up each object and other meta-info. After robot picking, those picked objects would be attached to robot end-effector, thus the planner must ensure picked objects are not in collision during subsequent robot movements.

As mentioned in Subsec.~\ref{subsec:example}, \texttt{MoveToPick} has only one major user-parameter: the \texttt{srvID} which identifies the perception service. Given the \texttt{srvID}, the perception message that contains objects and grasping information can be found in the variable map (environment), as shown in Table~\ref{table:table_vars}. This node also needs a \texttt{trajectory\_config} user-parameter. By default, an end-effector straight-line movement is used as \texttt{trajectory\_config}. Additionally, the user might specify a set of filters to specify additional requirements for the grasping candidates according to various factors, such as object type, picking pose, and the number of picked objects.

The behavior planning needs to select the grasping from a set of candidates. That includes selecting the object(s) to pick, and the end-effector pose for the selected object(s). An illustration is shown in Figure.~\ref{fig:decision} (a) and (b). Due to the symmetry of the objects and the gripper tool, there can be tens or hundreds of possible grasping poses for each object instance. Combined with tens of objects, the planner might need to attempt thousands of grasping candidates. 
The induced computation can be rather expensive, as the picking decision must be jointly made with subsequent movement nodes, as shown in Subsec.~\ref{subsec:example}. After making the picking decision, the planner also generates the robot trajectory that reaches the grasping pose. 

Given the picking decision and reaching robot joint-space trajectories as the online-parameters, the execution of \texttt{MoveToPick} node would: 1) remove objects that are selected for picking from the perception result variable, and insert them into \texttt{picked\_objects} variable (after updating attachment and meta-info); 2) execute the movement to reach the picking configuration in the same way as movement nodes.

It is emphasized that the \texttt{MoveToPick} node does not involve the \textit{physical} actions for robot picking behaviors. For example, picking up an object might require turning on the vacuum gripper or closing the parallel-jaw. To execute picking physically, nodes must be set up to send a control message to the gripper (which is an RPC service) or set a DigitalOut on the robot (if the gripper is connected to the robot). Thus, these physical nodes depend on hardware and connection configurations. Typically, the node for turning on the suction cups in a vacuum gripper is connected before \texttt{MoveToPick}, while node for closing the parallel-jaw is connected right after \texttt{MoveToPick}.

\begin{figure}[t]
    \centering
    \includegraphics[width=0.85\textwidth]{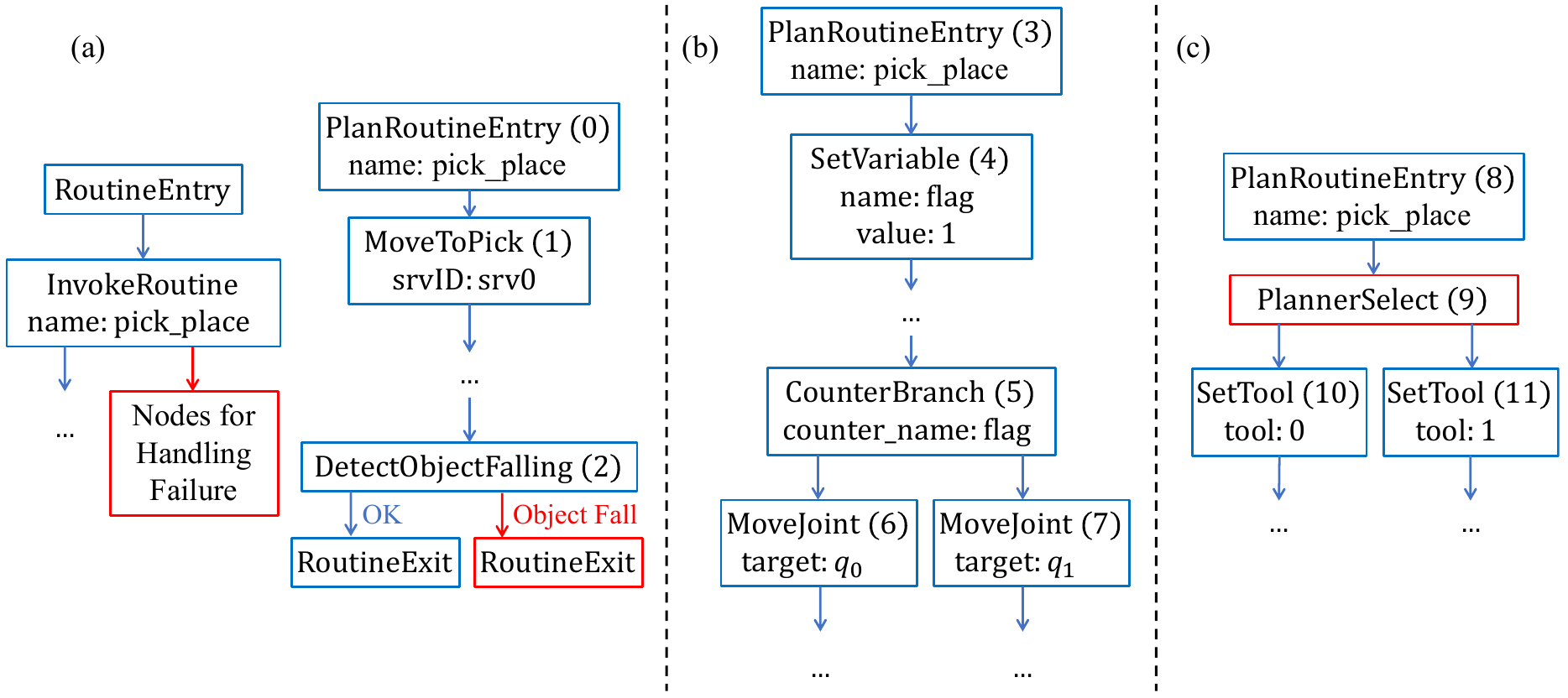}
    \caption{Branch structure types supported by a plan-routine. (a) The branch is explicitly marked as an exception. (b) The branch can be decided from data flow analysis. (c) The \texttt{PlannerSelect} branch used to set up decision-making problems regarding arbitrary factors, such as gripper tool selection in this example. Please refer to Subsec.~\ref{subsec:pr_struct} for a detailed explanation. }
    \label{fig:branch}
\end{figure}

\subsection{Structure of Plan Routines}
\label{subsec:pr_struct}

In Subsec.~\ref{subsec:example}, we discuss the plan-routine under the assumption that it is a simple sequence of nodes. In this sub-section, we relax this constraint by adding several types of branch nodes into the plan-routine, as illustrated in Figure.~\ref{fig:branch}. These types of branch nodes can be converted into a set of node sequences during planning, thus the planning algorithm in Subsec.~\ref{subsec:planning_alg} can be used to solve it. On the other hand, plan-routine cannot contain loops. This rule is enforced by static checking during the conversion from frontend to backend.

\vspace{1mm}
\noindent \textbf{Exception branch:} Some branch edges in a plan-routine are explicitly marked as \textit{exception} by the user. Usually, these exceptions are abnormal behaviors due to unmodeled effects. These exception edges are simply ignored during planning, while they might take effect in execution. An illustration is provided in Figure.~\ref{fig:branch} (a), the \texttt{DetectObjectFalling} node is used to check whether the objects have been successfully picked up (e.g., by force sensors on the gripper). If the picking is not successful, nodes for error recovery are executed.

Each plan-routine has a built-in \textit{PlanFailure} exception. \texttt{RoutineInvoke} nodes that invoke a plan-routine would have an out-port corresponding to this exception. Users might address the failure of behavior planning by various methods. For example, in Subsec.~\ref{subsec:demos} we use a vibration generator to create a random disturbance to objects in the container. Then, another image is taken and the planning is retried.

\vspace{1mm}
\noindent \textbf{Branch determined before planning:} Some branches can be determined during the construction of the planning problem. For example, in Figure.~\ref{fig:branch} the branch selection at \texttt{CounterBranch} node (5) can be determined from the \texttt{SetVariable} node (4). In general, determining the branch is a standard dataflow analysis problem, which is further simplified as plan-routines can not contain loops. A plan routine with this type of branch nodes can be converted into a node sequence, which is solved using the planning algorithm in Sec.~\ref{subsec:planning_alg}.

\vspace{1mm}
\noindent \textbf{Branch decided by the planner:} As shown in Figure.~\ref{fig:branch} (c), a \texttt{PlannerSelect} node (9) is used to request the planner to make a selection of gripper tools. This decision must be made jointly with other nodes in the plan-routine, similar to other discrete decisions in Figure.~\ref{fig:decision}. During planning, a plan-routine with \texttt{PlannerSelect} nodes would be expanded into a set of node sequences. The \texttt{PlannerSelect} node can be used to set up decision-making problems regarding arbitrary factors, such as object types, movement trajectories, and placement locations.

\section{Interpreter Implementation}
\label{sec:interpreter}

As mentioned in Sec.~\ref{sec:interface}, our DSL introduces plan-routines into the control flow graph. These plan-routines contain intentionally undetermined nodes that require online-parameters, such as movement trajectories and/or picking decisions. As a result, the interpreter of our DSL is responsible for calling the planner for these plan-routines when they are invoked during execution. Aside from that, the interpreter of our DSL behaves the same as the ones in existing programming languages: executing each node one by one, updating the variable map (environment), and generating side effects. In this section, we present the implementation of the interpreter. Subsec.~\ref{subsec:planning_alg} describes the planning algorithm. Subsec.~\ref{subsec:preplanning} presents a ``pre-planning'' mechanism to reduce the cycle-time of the manipulation pipeline.

\subsection{Planning Algorithm}
\label{subsec:planning_alg}

In this subsection, we present the algorithm that generates the online-parameters of the plan-routine. Our discussion is focused on the simple sequence, as branch structures presented in Subsec.~\ref{subsec:pr_struct} can be converted into a set of sequences. 

We use a specialization of the method in~\cite{garrett2018sampling} as the planner. In particular, we formulate a hybrid state transition system following~\cite{garrett2018sampling}. The state of this state transition system is a set of variables, such as ones in Table.~\ref{table:table_vars}. In addition to them, other problem-specific variables can also be included in the state, for example a \texttt{palletization\_state} variable is used to maintain the state (palletization pattern, packed boxes and next available slots) for palletization applications. The actions, which are converted from the nodes, define a set of constraints between the state before and after it. Using this formulation, the plan-routine becomes a set of ``skeletons'' described in~\cite{garrett2018sampling}. Then, a series of conditional samplers, which implement basic primitives (e.g., inverse kinematics, grasping pose sampling, and motion planning), are composed into a constraint sampling network (Figure. 8 of~\cite{garrett2018sampling}), which is used to generate online-parameters.

\cite{garrett2021integrated} also proposed algorithms that search for the skeleton (jointly with the online-parameters). As a result, many intelligent manipulation behaviors can emerge automatically, such as ``moving away the surrounding obstacles before reaching the target''. However, searching for the skeleton can be expensive due to the large solution space.
In this work, we take a different trade-off with more emphasize on computational performance: the skeletons are provided by the human operators through the DSL.
This approach aims to maximize the inherent advantages of the human operators and the planner. We rely on the planner to operate over the domain in which it outperforms the human, such as accurate and fast numerical computation, while leaving tasks that require cognition, such as high-level supervision, to the human operator.

\subsection{Pre-Planning Mechanism}
\label{subsec:preplanning}

The planning in Subsec.~\ref{subsec:planning_alg} can be time-consuming due to the large solution space and expensive operations (e.g., the collision detection). When the perceived scene and planning-routine are complex, the robot might need to stop the movement and wait for the planning result. To alleviate this issue, we propose to interleave the planning for future nodes with the execution. This is fruitful because we can exploit the time spent on waiting for RPC responses, which can be long from some services (e.g., robot service and perception service). For example, while executing the plan-routine for pick-place iteration 1, we would perform planning (in a background thread) for iteration 2 or 3. Thus, when executing pick-place iteration 2 the online-parameters of nodes are ready. This mechanism is referred as ``pre-planning'' in the following text.

\begin{figure}[t]
    \centering
    \includegraphics[width=0.3\textwidth]{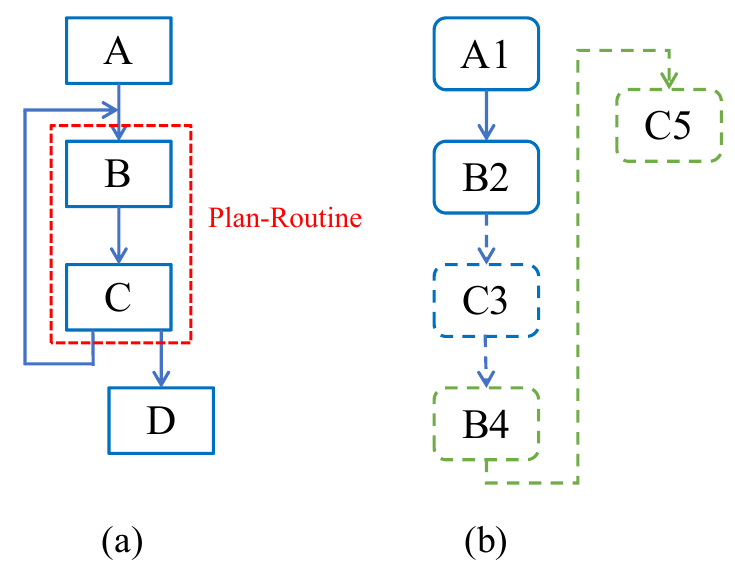}
    \caption{Example used to illustrate the pre-planning mechanism. For simplicity, we omit the routine structure and use the red dash-line block to represent nodes B and C are in a plan-routine. Please refer to Subsec.~\ref{subsec:preplanning} for a detailed explanation. }
    \label{fig:preplan}
\end{figure}

Consider the example in Figure.~\ref{fig:preplan}. For simplicity, we omit the routine structure and use the red dash-line block to imply nodes B and C are in a plan-routine. We assign a dynamic ID to each \textit{execution} of a node. Nodes with annotated dynamic ID is shown in Figure.~\ref{fig:preplan} (b). In the first loop iteration, the planner generate online parameters for (B2, C3). Suppose nodes A1 and B2 have been executed, and we would like to perform planning for the plan-routine in the next loop iteration, namely nodes (B4, C5).

To perform planning for (B4, C5), we need the variable map ``as if'' node C3 is executed. To achieve this, we implement a \texttt{simulate} interface for nodes in our DSL. This interface tries to update the variable map without generating side effect, and reports failure (which stops the pre-planning) if it is impossible. For several types of nodes, their \texttt{simulate} interface would:

\vspace{-3mm}
\noindent \textbf{CallService:} Mark the response variable as a special flag \texttt{AvailableUponExecution}. Reference to the response variable during subsequent \texttt{simulate} would report failure.

\vspace{-3mm}
\noindent \textbf{Movements:} Update the \texttt{jps} variable without sending request to robot service.

\vspace{-3mm}
\noindent \textbf{MoveToPick:} Update the \texttt{jps}, \texttt{picked\_objects} and \texttt{\{srvID\}\_perception} variables.

\vspace{-3mm}
\noindent \textbf{FunctorVariableMutation:} This node and nodes derived from it (e.g., \texttt{IncreaseCounter}) behave the same as execution, if no referred variable is marked as \texttt{AvailableUponExecution}.

\vspace{-3mm}
\noindent \textbf{ExceptionBranch:} Select the first branch that is not marked as exception by the user, such as ``OK'' branch in Figure.~\ref{fig:branch} (a).

\vspace{-3mm}
\noindent \textbf{OtherBranch:} Similar to \texttt{FunctorVariableMutation}.

Using the \texttt{simulate} interface, the interpreter would maintain another program counter and variable map for pre-planning. Then, the planning algorithm can be invoked (in a background thread) for future plan-routines using this variable map, and the planning results (online-parameters) can be directly used without waiting when these plan-routines are ready for execution. The pre-planning program counter and variable map would be reset to execution values, if 1) a node or the planner refers a variable marked as \texttt{AvailableUponExecution}; and 2) the guess in an exception branch node is wrong. This is similar to the pipelining and misprediction recovery in modern CPUs.

\section{Results}
\label{sec:result}

In this section, we first demonstrate a variety of industrially important applications that are implemented in our DSL, in Subsec.~\ref{subsec:demos}. These demonstration are achieved on several different hardwares regrading robot platforms, gripper tools, RGBD sensors and external devices. Then, we show the effectiveness of the proposed pre-planning mechanism in Subsec.~\ref{subsec:pre_demos}. These examples are illustrated in the accompanied video. Please visit \href{https://www.mech-mind.com/}{\underline{https://www.mech-mind.com/}} for more examples.

\subsection{Representative Examples}
\label{subsec:demos}

\begin{figure}[t]
    \centering
    \includegraphics[width=0.85\textwidth]{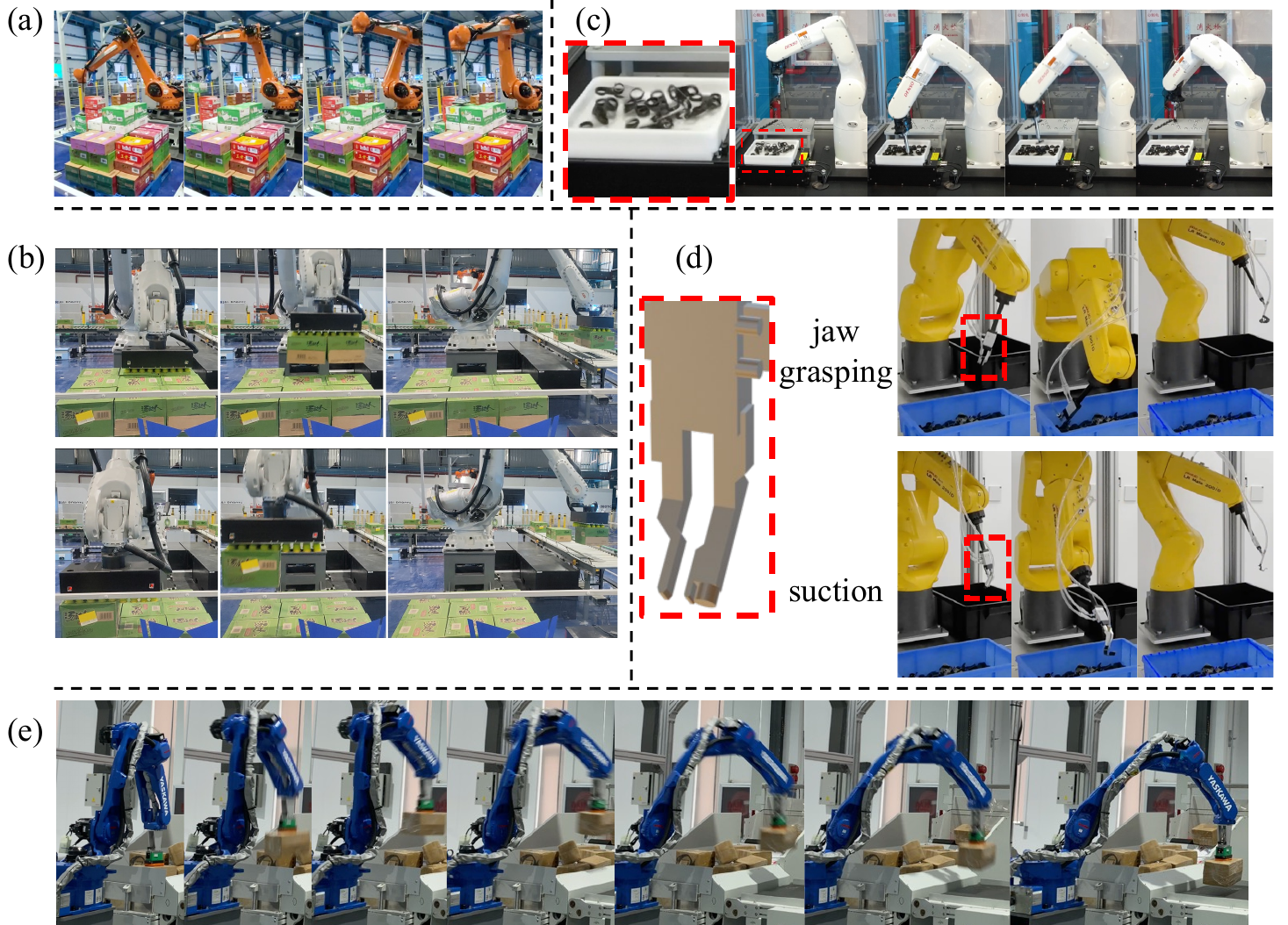}
    \caption{Representative industrial manipulation applications implemented with our DSL. (a) Mixed case palletization. (b) Multiple pick de-palletization. (c) Recovery from planning failure using exception (Subsec.~\ref{subsec:pr_struct}) and an external device (vibration generator). (d) Automatic selection of picking methods (suction or parallel-jaw grasping) for a dual-use gripper. (e) Integration with geometric motion planner. Please refer to Subsec.~\ref{subsec:demos} for a detailed description.}
    \label{fig:demo}
\end{figure}

\vspace{1mm}
\noindent \textbf{Mixed case palletization (a):} The robot performs a mixed-case (consisting of multiple types of boxes) palletization, as illustrated in Figure.~\ref{fig:demo} (a). The robot perceives the boxes using a camera and plans the robot actions that pick up a suitable box, transfer it and place it on the palletization. The decisions (e.g., selection of the box and the placement location) in this example must be made according to the desired palletization pattern that tries to maximize the space utilization rate.

\vspace{1mm}
\noindent \textbf{Multiple pick de-palletization (b):} The robot picks up boxes from a pallet and place them onto a conveyor, as shown in Figure.~\ref{fig:demo} (b). To improve the throughput (\# of boxes per hour), the robot might pick up multiple boxes at once. The pipeline detects currently available boxes using a camera, makes decisions about picking one or more boxes, and generates concrete picking behaviors and robot trajectories.

\vspace{1mm}
\noindent \textbf{Recovery from planning failure (c):} The robot picks workpieces and organizes them into a specific shape. During manipulation, the planner might fail to find a feasible pick-and-place behavior (e.g., the reaching movement collides with workpieces other than the picked one). The exception branch in Subsec.~\ref{subsec:pr_struct} is used to address the planning failure. In particular, an external vibration generator is used to create a random disturbance to these workpieces in the container, as shown in Figure.~\ref{fig:demo} (c). After that, the perception and behavior planning is re-tried.

\vspace{1mm}
\noindent \textbf{Selection of different gripper tools (d):} The robot is equipped with a special gripper tool that can pick up the object by parallel-jaw or air suction, as shown in Figure.~\ref{fig:demo} (d). These two types of grasping are treated as two logical gripper tools, and the pipeline automatically determines which one to use. As shown in Figure.~\ref{fig:branch}, this decision should be made incorporating the nodes for reaching, picking, transferring and placement. During execution, the digital output corresponding to either closing the parallel jaw or turning on the suction cup is invoked to pick up the objects.

\vspace{1mm}
\noindent \textbf{Integration of geometric motion planner (e):} Our DSL provides a flexible interface for the integration of collision-free motion planners, such as sampling~\cite{kavraki1996probabilistic,lavalle2001randomized} and optimization~\cite{ratliff2009chomp,schulman2014motion} based methods, as the \textit{motion generation} primitives of the TAMP planner in Subsec.~\ref{subsec:planning_alg}. These collision-free motion planners are accessed through the \texttt{trajectory\_config} user-parameter, as detailed in Subsec.~\ref{subsec:move_nodes}. An example is shown in Figure.~\ref{fig:demo} (e), the shortcut algorithm in~\cite{geraerts2007creating} is used to generate a smooth and efficient robot transferring movement of the picked object.

\subsection{Effectiveness of the Pre-Planning Mechanism}
\label{subsec:pre_demos}

\begin{figure}[t]
    \centering
    \includegraphics[width=0.6\textwidth]{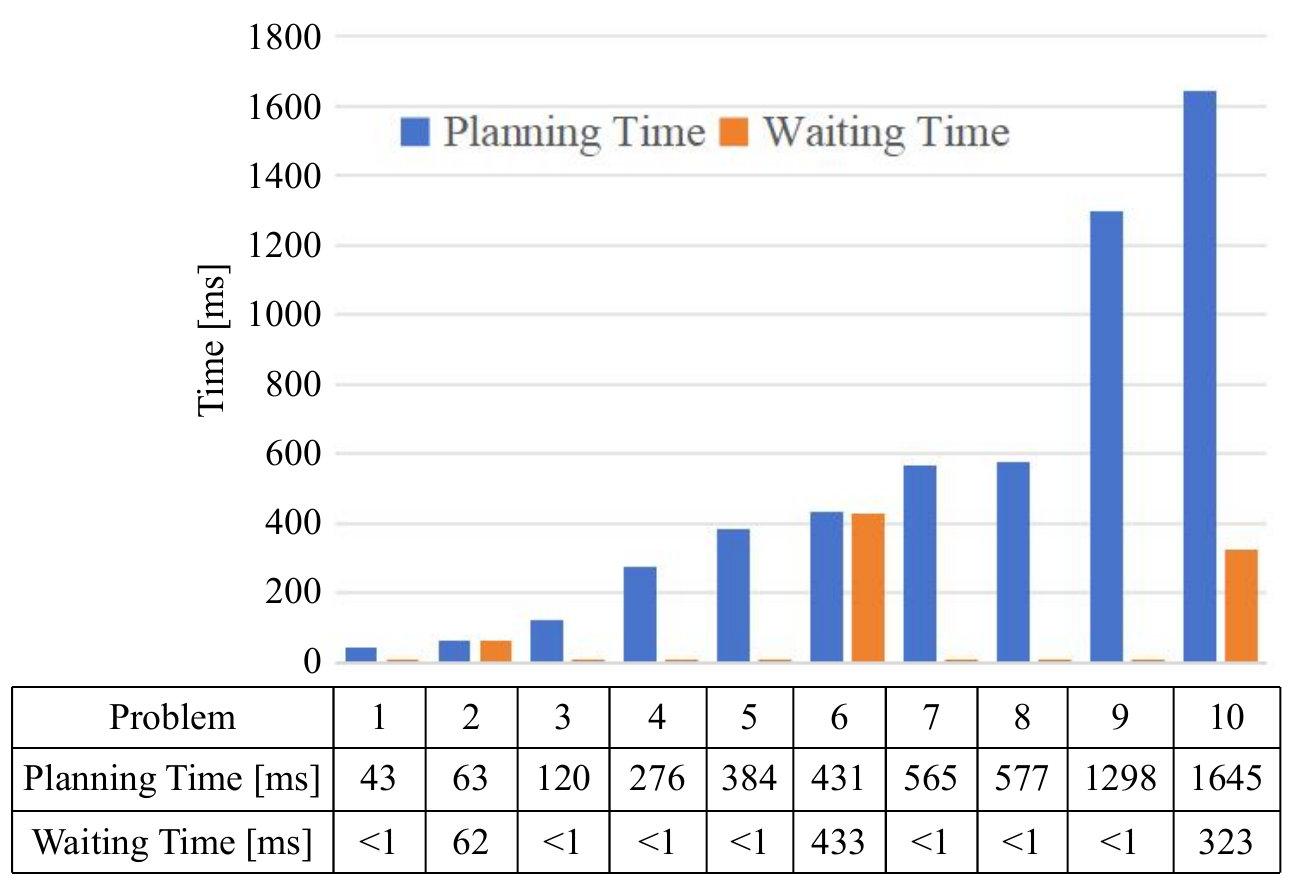}
    \caption{The planning time and waiting time for 10 planning problems. With the proposed pre-planning mechanism in Subsec.~\ref{subsec:preplanning}, the time that the robot must be stopped for waiting the planning result can be eliminated or significantly reduced in many cases. Please refer to Subsec.~\ref{subsec:pre_demos} for more details.}
    \label{fig:preplanning_time}
\end{figure}

The pre-planning mechanism proposed in Subsec.~\ref{subsec:preplanning} is used to interleave the node execution (e.g., waiting for RPC responses from robot services) with planning for future plan-routines. To illustrate its effectiveness, we compare the planning time with the time that the interpreter spends on waiting for the planning result. If the pre-planning mechanism successfully exploits the node execution time for planning, the waiting time should be much shorter than the planning time. The results are shown in Figure.~\ref{fig:preplanning_time} for 10 different planning problems in user programs. Among them, 5 planning problems are from the examples in Subsec.~\ref{subsec:demos}. Each planning problem is invoked 5-30 times in the user program, and the times are the average of 20 runs of the user program.

From the result, in most cases the pre-planning mechanism can eliminate or significantly reduce the waiting time. Thus, the robot does not need to stop and waiting for online-parameters before execution. On the other hand, the pre-planning can not help when the required variables are not ready (problems 2 and 6). Moreover, if the planning time is very long (problem 10), the waiting time can not be fully eliminated.

\section{Limitations and Future Works}
\label{sec:limitations}

Currently, the DSL mainly focuses on executing planned trajectories in an open-loop way. Reactive, closed-loop control (e.g., visual servoing) is not supported in our DSL. Moreover, our DSL assumes that the manipulated objects are (mostly) rigid. This abstraction does not work for deformable objects or more dexterous manipulation actions on rigid objects, such as the in-hand manipulation in~\cite{andrychowicz2018learning}. Deploying these interesting manipulation skills into industrial production at scale is still challenging, and it is a promising direction for future work.

In terms of the implementation, our DSL evolved from a GUI application to simplify the deployment of the perception-guided manipulation pipeline. During the early stage of development, many concepts are unclear as we have not realized the software should be designed as a programming language, and various shortsighted design decisions have been made. The graphical user interface designed at that stage, which was already used by many customers, became historical baggage. 
Thus, several features of the DSL can only be provided in an incomplete and/or unnatural way. We are re-factorizing our code base to address this issue.

Currently, user programs in our DSL are crafted by human operators. One approach to further simplify the deployment is to train Large Language Models (LLM) to generate code in our DSL. This might be fruitful as our DSL is mainly used by field application specialists without coding experience in traditional programming languages.

\section{Conclusion}
\label{sec:conclusion}

This paper contributes a DSL for deploying perception-guided robotic manipulation at scale. This DSL has an intuitive graphical frontend and is mainly used by machine operators without coding experience in Python/C++. To alleviate the users from manually making discrete decisions and/or crafting robot trajectories, we propose a novel interface that integrates a TAMP algorithm into the DSL. In particular, users craft a ``skeleton'' of the desired robot behavior with a set of intentionally undetermined parameters, and the planning algorithm automatically generates these missing parameters during online execution. With this interface, users can setup and solve practically important TAMP problems without understanding the TAMP algorithm or explicitly writing the planning problem description (in a modeling language like PDDL). Moreover, we propose a pre-planning interpreter to reduce the cycle time and improve the throughput of the manipulation applications. Extensive practical applications in industry demonstrate the efficacy of our method.

\subsubsection*{Acknowledgments}

The authors would like to thank Xi Li and Lili Yang for their insightful discussion and maintenance of the infrastructure code. This work was conducted during the authors' employment at Mech-Mind Robotics. The views expressed in this paper are those of the authors themselves and are not endorsed by the supporting agencies.

\bibliographystyle{apalike}
\bibliography{main.bib}

\end{document}